\documentclass[letterpaper, 10 pt, conference]{ieeeconf}
\IEEEoverridecommandlockouts
\usepackage{hyperref}
\hypersetup{hidelinks,
	colorlinks=true,
	allcolors=black,
	pdfstartview=Fit,
	breaklinks=true}
\usepackage{algorithmic}
\usepackage{algorithm}
\usepackage{graphicx}
\usepackage{amsmath}
\usepackage{gensymb} 
\usepackage[caption=false,font=footnotesize,labelfont=rm,textfont=rm]{subfig}
\usepackage{booktabs}
\usepackage{threeparttable}

\title{\LARGE \bf
High-Precision and High-Efficiency Trajectory Tracking for Excavators Based on Closed-Loop Dynamics
}

\author{Ziqing Zou$^{1}$, Cong Wang$^{2}$, Yue Hu$^{2}$, Xiao Liu$^{2}$, Bowen Xu$^{2}$, Rong Xiong$^{1}$,\\Changjie Fan$^{2}$, Yingfeng Chen$^{2, *}$ and Yue Wang$^{1, *}$
\thanks{$^{1}$Zhejiang University.
$^{2}$Fuxi Robotics Lab, NetEase Inc.}
\thanks{
$^{*}$Corresponding authors: Yingfeng Chen and Yue Wang.}
\thanks{
This work was supported by CCF-NetEase ThunderFire Innovation Research Funding under Grant 202305.}
}

\begin{document}

\maketitle
\thispagestyle{empty}
\pagestyle{empty}

\begin{abstract}
The complex nonlinear dynamics of hydraulic excavators, such as time delays and control coupling, pose significant challenges to achieving high-precision trajectory tracking. Traditional control methods often fall short in such applications due to their inability to effectively handle these nonlinearities, while commonly used learning-based methods require extensive interactions with the environment, leading to inefficiency. To address these issues, we introduce EfficientTrack, a trajectory tracking method that integrates model-based learning to manage nonlinear dynamics and leverages closed-loop dynamics to improve learning efficiency, ultimately minimizing tracking errors. We validate our method through comprehensive experiments both in simulation and on a real-world excavator. Comparative experiments in simulation demonstrate that our method outperforms existing learning-based approaches, achieving the highest tracking precision and smoothness with the fewest interactions. Real-world experiments further show that our method remains effective under load conditions and possesses the ability for continual learning, highlighting its practical applicability. For implementation details and source code, please refer to \href{https://github.com/ZiqingZou/EfficientTrack}{https://github.com/ZiqingZou/EfficientTrack}.
\end{abstract}

\begin{figure*}[!t]
\vspace{5pt}
\centering
\includegraphics[width=6.9in]{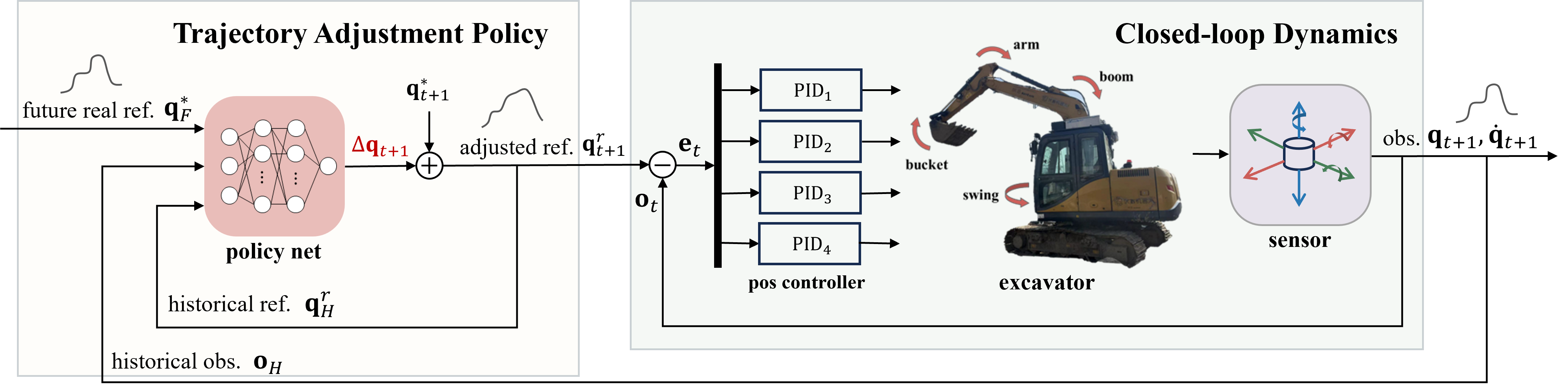}
\vspace{-5pt}
\caption{Control block diagram of our method. We integrate a trajectory adjustment policy into the excavator's closed-loop dynamics. During implementation, the policy adjusts reference trajectories in real time to preemptively compensate for control errors, guiding the control system to achieve high-precision trajectory tracking.}
\vspace{-12pt}
\label{control_block}
\end{figure*}

\section{Introduction}

Excavators are primarily used in earthworks, mining, and construction projects, playing a vital role in tasks such as digging, loading, trenching, and leveling \cite{shibanov2021digital, huo2023intelligent, fu2023digital}. Automation in these processes not only increases efficiency, but also reduces costs and enhances safety. The complete automation workflow for excavator operations generally includes environmental perception, high-level decision making, motion planning, and trajectory tracking \cite{zhang2021autonomous, eraliev2022sensing}. Among these components, the precision of trajectory tracking is crucial. Improved tracking performance often leads to more flexible and sophisticated operational capabilities.

Typically, trajectory tracking for dynamic systems is modeled as an optimal control problem aimed at minimizing tracking errors.
Hydraulic excavators, as multi-input multi-output (MIMO) systems, present complex nonlinear dynamics that pose substantial challenges to their optimal control.
In addition to the nonlinear forces of inertia and gravitation, excavators also face control coupling, state-dependent dead zones, and time delays caused by hydraulic drives \cite{ref5, ref14}.

Optimal control of trajectory tracking is usually achieved with a controller, which is always designed based on state feedback from sensors. 
Traditional control methods have long sought to find the optimal controller for dynamic systems but often require cumbersome mechanistic modeling or otherwise struggle to cope with the complex nonlinearities.
In contrast, learning-based methods with neural networks are well-suited for handling nonlinear dynamics due to their inherent nonlinear fitting and prediction capabilities.
Reinforcement Learning (RL) algorithms optimize long-term rewards through interactions with the environment, achieving optimal control. However, they suffer from inherent variance in the estimation of future returns via value functions, limiting their precision in trajectory tracking \cite{ref18,ref19,ref17}.
Model-based learning encapsulates the system's state transitions into a world model to predict future states. Backpropagation through the world model bypasses the estimation of the value function and helps to minimize long-term tracking errors \cite{ref20}. However, learning an accurate world model requires extensive exploration in the environment, leading to inefficiency.

In this paper, we propose an innovative framework that integrates closed-loop dynamics into model-based learning, where the system dynamics shift from global dynamics across the entire state space to local dynamics around the reference trajectories, as shown in Fig. \ref{control_block}.
Within our proposed framework, the reference trajectories provide an initial guess for the closed-loop dynamics model's predictions, allowing the model to focus solely on capturing residual tracking errors. This approach avoids the need to learn complete state transitions from scratch, resulting in high learning efficiency.
Subsequently, we optimize a reference trajectory adjustment policy based on this prediction model. It effectively applies a correction term outside the closed-loop dynamics to guide the controller's actual tracking performance closer to the real reference, compensating for residual tracking errors.

Our main contributions are summarized as follows:
\begin{itemize}
\item We propose a framework that learns the closed-loop dynamics of excavators, bypassing the need to model their complex nonlinear dynamics, thereby improving learning efficiency.
\item We apply multi-step gradient backpropagation to both the closed-loop dynamics model and the trajectory adjustment policy. By directly optimizing multi-step model prediction errors and policy tracking errors, we enhance tracking precision.
\item Comparative experiments in simulation demonstrate that our method achieves the smallest tracking errors with the fewest interactions. Real-world excavator experiments further validate the effectiveness of our method.
\end{itemize}

\section{Related Work}
\subsection{Optimal Control of Excavators}
\subsubsection{Traditional Control}
Model Predictive Control (MPC) is widely utilized for solving optimal control problems. Offset-free MPC \cite{ref1} is employed using the accelerated proximal gradient method to solve the underlying optimization problem, achieving steady-state tracking on a 1/16 hydraulic mini excavator. Similarly, nonlinear MPC \cite{ref2} is used to complete the loading task of a 20-ton excavator in simulation, achieving both position control and obstacle avoidance.
To enhance the disturbance rejection capability of control systems, Sliding Mode Control (SMC) has also been applied to excavator control. Super twisting nonsingular terminal SMC \cite{ref3} is used to achieve high-precision trajectory tracking on a motor-driven 1/35 scale excavator. Furthermore, fuzzy adaptive SMC \cite{ref4} is employed to achieve centimeter-level trajectory tracking for leveling operations on a hydraulic excavator.
These traditional control methods require precise modeling of the excavator's forward dynamics and hydraulic drive system, which adds extra cost and hinders generalization.
Proportional-Integral-Derivative (PID) control, while not requiring mechanistic modeling, typically does not handle complex nonlinearity well and is often used as part of a larger control system to enhance control performance \cite{ref10}.

\subsubsection{Learning-Based Methods}
Learning-based methods do not rely on mechanistic modeling or parameter tuning but instead seek optimal control policies from data.
Nonlinear echo-state networks (ESNs) \cite{ref11} have been used to learn an inverse model, achieving performance comparable to PD controllers for tracking planar trajectories of excavators. A modular design, incorporating an infinite impulse response unit to accommodate input delays, a piecewise linear map to deal with dead-zones, and Multi-Layer Perceptron (MLP) networks to capture the remaining nonlinear dynamics \cite{ref21}, is employed for precise motion control in digging and grading tasks, though it requires large amounts of training data. In \cite{ref12}, a heteroscedastic Gaussian Processes (GPs) prediction model is utilized to minimize the designed cost through gradient descent, for phase switching and trajectory generation to capture a desired amount of soil. However, its optimal performance susceptible to the curse of dimensionality, making it relies on simple parameterization of the action space.
The urgent challenge for learning-based methods is to achieve better control performance with less training data.

\subsection{Reinforcement Learning}
Reinforcement Learning (RL) aims to find policies that maximize rewards through interaction with the environment and can be used to solve optimal control problems for excavators.
Proximal Policy Optimization (PPO) \cite{ref18} stabilizes the training process by limiting the extent of policy updates. It has few hyperparameters and is easy to use, making it one of the first algorithms to consider for decision and control problems in robotics. In \cite{ref14}, \cite{ref15}, and \cite{ref7}, PPO is used to achieve end-effector position tracking, end-effector velocity tracking of circular trajectories, and soil-adaptive digging, respectively. In \cite{ref16}, PPO is used for autonomous loading of excavators in simulation. Additionally, Soft Actor-Critic (SAC) \cite{ref19} enhances exploration by maximizing the entropy of the policy, making it another commonly used model-free RL algorithm.
Model-free RL has shown good performance in decision-making and control tasks, but it requires extensive exploration in the environment, leading to inefficiency.

Temporal Difference Learning for Model Predictive Control (TD-MPC) \cite{ref17} is a state-of-the-art model-based RL algorithm that encodes real-world observations into a latent space, using consistency loss to ensure latent variables contain sufficient historical information and the model maintains strong predictive capabilities. SuperTrack \cite{ref20} is also a model-based learning method developed for motion tracking of physically simulated characters, utilizing multi-step gradient backpropagation to minimize multi-step errors. Similarly, our method leverages this training approach to enhance tracking performance. However, SuperTrack's world model learns by rolling out the policy in the environment, and the exploration of redundant state spaces makes its learning very inefficient. Our method addresses this issue by separating model training and policy optimization through an effective data sampling approach (see \ref{data collection}).

\section{Problem Statement}
In this section, we model the trajectory tracking of the excavator's closed-loop dynamics as a Partially Observable Markov Decision Process (POMDP) \cite{lauri2022partially} and formulate the corresponding optimal control problem.

Sensors have been installed at the excavator's joints, so the observation of the closed-loop dynamics system is defined by the angular positions and velocities of the excavator's four joints: boom, arm, bucket and the cab's swing, as shown in Fig.\ref{control_block}. Let \( q_t^i \) and \( \dot{q}_t^i \) represent the position and velocity of joint \( i \) at time \( t \). The position vector is:
\begin{equation}
\setlength\abovedisplayskip{5pt}
\setlength\belowdisplayskip{4pt}
\mathbf{q}_t = \left( q_t^1, q_t^2, q_t^3, q_t^4 \right)^T 
\end{equation}
and the observation vector is:
\begin{equation}
\mathbf{o}_t = \left( q_t^1, q_t^2, q_t^3, q_t^4, \dot{q}_t^1, \dot{q}_t^2, \dot{q}_t^3, \dot{q}_t^4 \right)^T 
\end{equation}

To provide sufficient information for policy inference, we employ a closed-loop dynamics model \(g_\theta\) that incorporates historical observations and reference positions to predict the change in observations:
\begin{equation}
\setlength\abovedisplayskip{5pt}
\setlength\belowdisplayskip{5pt}
\mathbf{\hat{o}}_{t+1} = \mathbf{o}_t + g_\theta\left(\mathbf{o}_H, \mathbf{q}^r_H, \mathbf{q}^r_{t+1}\right)
\label{model equation}
\end{equation}
where the historical observations are \(\mathbf{o}_H = \left(\mathbf{o}_{t-h}, ..., \mathbf{o}_t\right)^T\) and the historical references are \(\mathbf{q}^r_H = \left(\mathbf{q}^r_{t-h}, ..., \mathbf{q}^r_{t}\right)^T\).
When \(t<h\), the terms preceding \(\mathbf{o}_0\) and \(\mathbf{q}^r_0\) are padded with \(\mathbf{o}_0\) to maintain consistent input dimensions.

To further enhance tracking performance, we introduce a trajectory adjustment policy \(\pi_\phi\) that outputs adjustment actions based on the historical observations \(\mathbf{o}_H\), historical reference positions \(\mathbf{q}^r_H\), and future desired references \(\mathbf{q}^*_F = \left(\mathbf{q}^*_{t+1}, ..., \mathbf{q}^*_{t+h}\right)^T\). Mathematically, the policy is defined as:
\begin{equation}
\setlength\abovedisplayskip{5pt}
\setlength\belowdisplayskip{5pt}
\mathbf{a}_t = \pi_\phi\left(\mathbf{o}_H, \mathbf{q}^r_H, \mathbf{q}^*_F\right)
\label{policy equation}
\end{equation}

We adjust the reference trajectories that the closed-loop system tracks, defining the action \( \mathbf{a}_t \) as the adjustment added to the desired position \( \mathbf{q}^*_{t+1} \). After applying this action, the reference position at time \( t+1 \) becomes:
\begin{equation}
\label{policy_action}
\setlength\abovedisplayskip{4pt}
\setlength\belowdisplayskip{5pt}
\mathbf{q}^r_{t+1} = \mathbf{q}^*_{t+1} + \mathbf{a}_t
\end{equation}

Ultimately, the optimal trajectory tracking control problem we aim to solve for the excavator's closed-loop dynamics can be stated as follows: Given the desired trajectory \(\mathbf{q}^*_{1:T}\) and the closed-loop dynamics \(g_\theta\), we seek an optimal trajectory adjustment policy \(\pi_\phi\) that iteratively adjusts the reference positions to minimize the position tracking error over a horizon of \(h\) steps:
\begin{equation}
\setlength\abovedisplayskip{0pt}
\setlength\belowdisplayskip{5pt}
\label{tracking_error}
J = \sum_{i=t}^{t+h} \| \mathbf{\hat{q}}_i - \mathbf{q}^*_i \|^2
\end{equation}

\begin{algorithm}[!t]
\caption{EfficientTrack}\label{algorithm1}
\begin{algorithmic}[1]
\STATE \textsc{CollectData}(\(\mathcal{S}_{reftraj},  \sigma_{max}\))

\STATE \hspace{0.4cm}\(\mathcal{S}_{dataset}\gets[]\)

\STATE \hspace{0.4cm}\textbf{for} \(reftraj\) \textit{\textbf{in}} \(\mathcal{S}_{reftraj}\) \textbf{do}

\STATE \hspace{0.8cm}\(\{\mathbf{q}^*_0, \mathbf{q}^*_1, ..., \mathbf{q}^*_{T}\} \gets reftraj\)

\STATE \hspace{0.8cm}\(\sigma\gets\) \text{random}(\(0, \sigma_{max}\))

\STATE \hspace{0.8cm}\(\textsc{EnvReset}(\mathbf{q}^*_0), \mathbf{q}^r_0\gets\mathbf{q}^*_0\)

\STATE \hspace{0.8cm}\textbf{for} \(t\) \textit{\textbf{in}} \(\text{range}(T)\) \textbf{do}

\STATE \hspace{1.2cm}\(\mathbf{o}_t\gets\textsc{GetObservation}(), (\mathbf{q}_t, \mathbf{\dot{q}}_t)\gets\mathbf{o}_t\)

\STATE \hspace{1.2cm}\(\mathbf{d}_t \sim \mathcal{N}(0, \sigma), \mathbf{d}_t \gets \text{clip}(\mathbf{d}_t, -\sigma, \sigma)\)

\STATE \hspace{1.2cm}\(\mathbf{q}^r_{t+1}\gets\mathbf{q}^*_{t+1}+\mathbf{d}_t \)

\STATE \hspace{1.2cm}\(\mathbf{u}_t\gets\textsc{PdController}(\mathbf{q}_t,\mathbf{q}_{t-1}, \mathbf{q}^r_{t+1},\mathbf{q}^r_t)\)

\STATE \hspace{1.2cm}\(done\gets\textsc{EnvStep}(\mathbf{u}_t)\)

\STATE \hspace{1.2cm}\begin{small}\(\mathcal{S}_{dataset}\gets\mathcal{S}_{dataset} \cup \{\mathbf{o}_t, \mathbf{q}^r_{t+1}, \mathbf{q}^*_{t+1}, \mathbf{u}_t, done\}\)\end{small}

\STATE \hspace{0.4cm}\textbf{return} \(\mathcal{S}_{dataset}\)

\STATE

\STATE \textsc{TrainModel}(\(\mathcal{S}_{dataset}, N_{epoch}, h, w\))

\STATE \hspace{0.4cm}\(g_\theta\gets\textsc{InitModel}()\)

\STATE \hspace{0.4cm}\textbf{for} \(\_\) \textit{\textbf{in}} \(\text{range}(N_{epoch})\) \textbf{do}

\STATE \hspace{0.8cm}\(idxs\gets \text{shuffle}([0, ..., \text{len}(\mathcal{S}_{dataset})])\)

\STATE \hspace{0.8cm}\textbf{for} \(t\) \textit{\textbf{in}} \(idxs\) \textbf{do}

\STATE \hspace{1.2cm}\resizebox{0.79\hsize}{!}{\(\{\mathbf{o}_{t-h}, ..., \mathbf{o}_{t+h}, \mathbf{q}^r_{t-h}, ..., \mathbf{q}^r_{t+h}\}\gets\textsc{Sample}(\mathcal{S}_{dataset})\)}

\STATE \hspace{1.2cm}\(\mathbf{o}_H\gets[\mathbf{o}_{t-h}, ..., \mathbf{o}_{t}], \hat{\mathbf{o}}_t\gets\mathbf{o}_t\)

\STATE \hspace{1.2cm}\(\mathbf{q}^r_H\gets[\mathbf{q}^r_{t-h}, ..., \mathbf{q}^r_t]\)

\STATE \hspace{1.2cm}\textbf{for} \(i\) \textit{\textbf{in}} \(\text{range}(t, t+h)\) \textbf{do}

\STATE \hspace{1.6cm}\(\hat{\mathbf{o}}_{i+1}\gets\hat{\mathbf{o}}_i+g_\theta(\mathbf{o}_H, \mathbf{q}^r_H, \mathbf{q}^r_{i+1})\)

\STATE \hspace{1.6cm}\(\mathbf{o}_H\gets\left[\mathbf{o}_H\text{[1:]}, \hat{\mathbf{o}}_{i+1}\right]\)

\STATE \hspace{1.6cm}\(\mathbf{q}^r_H\gets\left[\mathbf{q}^r_H\text{[1:]}, \mathbf{q}^r_{i+1}\right]\)

\STATE \hspace{1.2cm}\(\theta\gets\theta-\nabla_\theta\left(\sum_{i=1}^{h} \| \hat{\mathbf{o}}_{t+i} - \mathbf{o}_{t+i}\|^2 - w\|\theta\|^2\right)\)

\STATE \hspace{0.4cm}\textbf{return} \(g_\theta\)

\STATE

\STATE \textsc{TrainPolicy}(\(\mathcal{S}_{dataset}, N_{epoch}, h, g_\theta, \gamma, k_{smooth}\))

\STATE \hspace{0.4cm}\(\pi_\phi\gets\textsc{InitPolicy}()\)

\STATE \hspace{0.4cm}\textbf{for} \(\_\) \textit{\textbf{in}} \(\text{range}(N_{epoch})\) \textbf{do}

\STATE \hspace{0.8cm}\(idxs\gets \text{shuffle}([0, ..., \text{len}(\mathcal{S}_{dataset})])\)

\STATE \hspace{0.8cm}\textbf{for} \(t\) \textit{\textbf{in}} \(idxs\) \textbf{do}

\STATE \hspace{1.2cm}\resizebox{0.79\hsize}{!}{\(\{\mathbf{o}_{t-h}, ..., \mathbf{o}_{t+h}, \mathbf{q}^r_{t-h}, ..., \mathbf{q}^r_t\}\gets\textsc{Sample}(\mathcal{S}_{dataset})\)}

\STATE \hspace{1.2cm}{\(\{\mathbf{q}^*_{t+1}, ..., , \mathbf{q}^*_{t+2h}\}\gets\textsc{SampleRef}(\mathcal{S}_{dataset})\)}

\STATE \hspace{1.2cm}\(\mathbf{o}_H\gets[\mathbf{o}_{t-h}, ..., \mathbf{o}_{t}], \hat{\mathbf{o}}_t\gets\mathbf{o}_t\)

\STATE \hspace{1.2cm}\(\mathbf{q}^r_H\gets[\mathbf{q}^r_{t-h}, ..., \mathbf{q}^r_t], \mathbf{q}^*_F\gets[\mathbf{q}^*_{t+1}, ..., \mathbf{q}^*_{t+h}]\)

\STATE \hspace{1.2cm}\textbf{for} \(i\) \textit{\textbf{in}} \(\text{range}(t, t+h)\) \textbf{do}

\STATE \hspace{1.6cm}\(\mathbf{a}_i\gets\pi_\phi(\mathbf{o}_H, \mathbf{q}^r_H, \mathbf{q}^*_F), \mathbf{q}^r_{i+1}\gets\mathbf{q}^*_{i+1}+\mathbf{a}_i\)

\STATE \hspace{1.6cm}\(\hat{\mathbf{o}}_{i+1}\gets\hat{\mathbf{o}}_i+g_\theta(\mathbf{o}_H, \mathbf{q}^r_H, \mathbf{q}^r_{i+1})\)

\STATE \hspace{1.6cm}\(\mathbf{o}_H\gets\left[\mathbf{o}_H\text{[1:]}, \hat{\mathbf{o}}_{i+1}\right], \mathbf{q}^r_H\gets\left[\mathbf{q}^r_H\text{[1:]}, \mathbf{q}^r_{i+1}\right]\)

\STATE \hspace{1.6cm}\(\mathbf{q}^*_F\gets\left[\mathbf{q}^*_F\text{[1:]}, \mathbf{q}^*_{i+h+1}\right]\)

\STATE \hspace{1.2cm}\(\phi\gets\phi-\nabla_\phi\sum_{i=1}^{h}\gamma^i\| \hat{\mathbf{q}}_{t+i} - \mathbf{q}^*_{t+i}\|^2\)

\STATE \hspace{1.2cm}\(\phi\gets\phi-k_{smooth}\nabla_\phi\sum_{i=2}^{h}\|\mathbf{a}_{t+i}-\mathbf{a}_{t+i-1}\|^2\)

\STATE \hspace{0.4cm}\textbf{return} \(\pi_\phi\)

\end{algorithmic}
\end{algorithm}

\section{Method}
In this section, we provide a detailed explanation of how to train the closed-loop dynamics model (\ref{model}) to learn the observation transitions, and how to optimize the trajectory adjustment policy (\ref{policy}) to solve the optimal control problem. Algorithm \ref{algorithm1} provides details of our method.

\subsection{Closed-Loop Dynamics Model}
\label{model}
We utilize the closed-loop dynamics model \(g_\theta\) defined in equation (\ref{model equation}) to approximate the observation transitions of the excavator.

\subsubsection{Forward Propagation}
Given the future reference trajectory \( \mathbf{q}^r_F = \left(\mathbf{q}^r_{t+1}, ..., \mathbf{q}^r_{t+h}\right)^T \), the model can iteratively generate predictions through forward propagation. At each time step, the model predicts the next observation by incorporating historical observations and reference positions. The historical observations include real data from the dataset and the model's previously predicted observations.

\subsubsection{Backward Propagation}
To ensure accurate predictions across multiple time steps, we train the model parameters \(\theta\) using supervised learning with multi-step gradient backpropagation \cite{ref20}.
The loss function is designed to minimize the mean squared prediction error over the prediction horizon \(h\): 
\begin{equation}
\label{prediction_loss}
\setlength\abovedisplayskip{2pt}
\setlength\belowdisplayskip{4pt}
L(\theta) = \frac{1}{h} {\sum_{i=1}^{h} \| \hat{\mathbf{o}}_{t+i} - \mathbf{o}_{t+i}\|^2} + w\|\theta\|^2
\end{equation}
Here, \( \hat{\mathbf{o}}_{t+i} \) is the predicted observation from the model, and \( \mathbf{o}_{t+i} \) is the true observation from the dataset, both driven by the same reference sequence \(\mathbf{q}^r_F\). The term \( w\|\theta\|^2 \) is a weight decay regularization to prevent overfitting, with \(w\) being the regularization coefficient.

\begin{figure}[!t]
\centering
\includegraphics[width=3.4in]{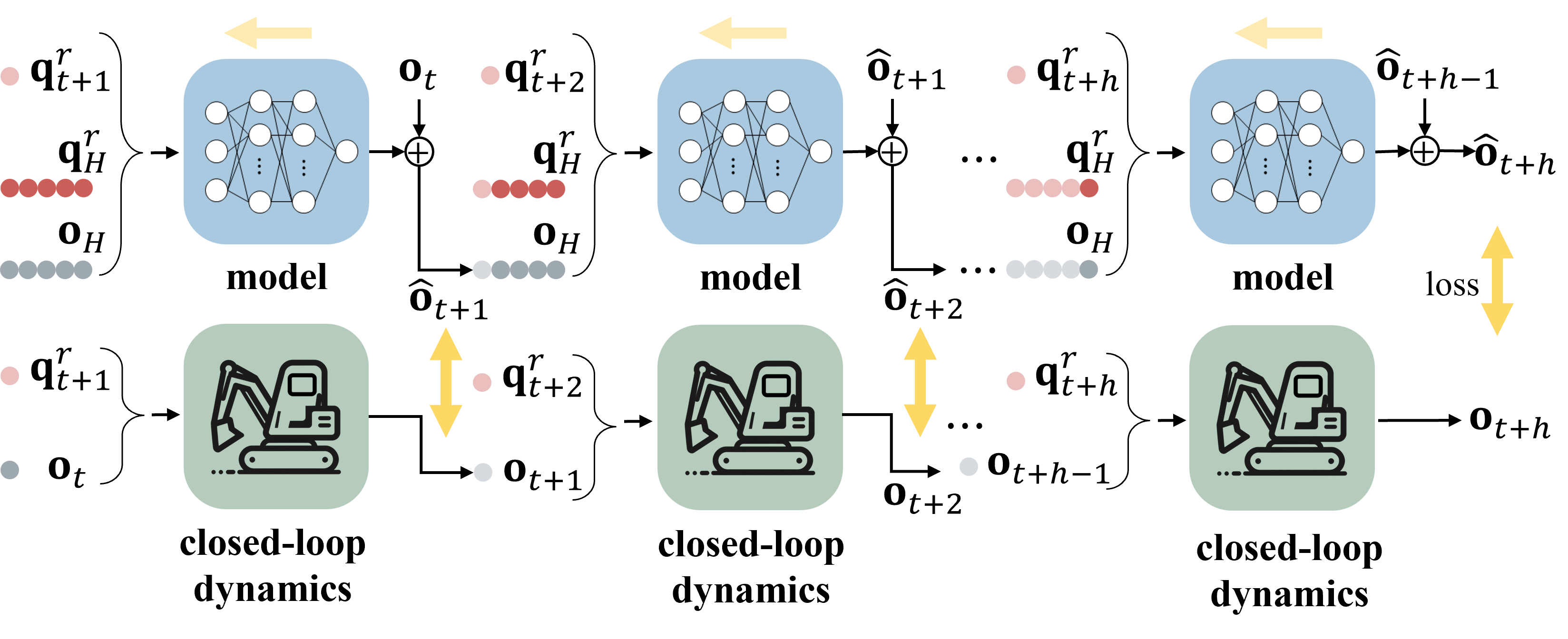}
\vspace{-18pt}
\caption{Multi-step forward and backward propagation of the observation predition model. The black arrows indicate the forward propagation of observations over time steps, while the yellow arrows indicate the backpropagation of the prediction loss starting from step \(t + i\) to step \(t\). Both the model and the closed-loop dynamics of the excavator are driven by the same reference sequence.}
\label{model_learning}
\vspace{-2pt}
\end{figure}

The multi-step forward and backward propagation processes of the model are shown in Fig. \ref{model_learning}.

\begin{figure}[!t]
\centering
\includegraphics[width=3.2in]{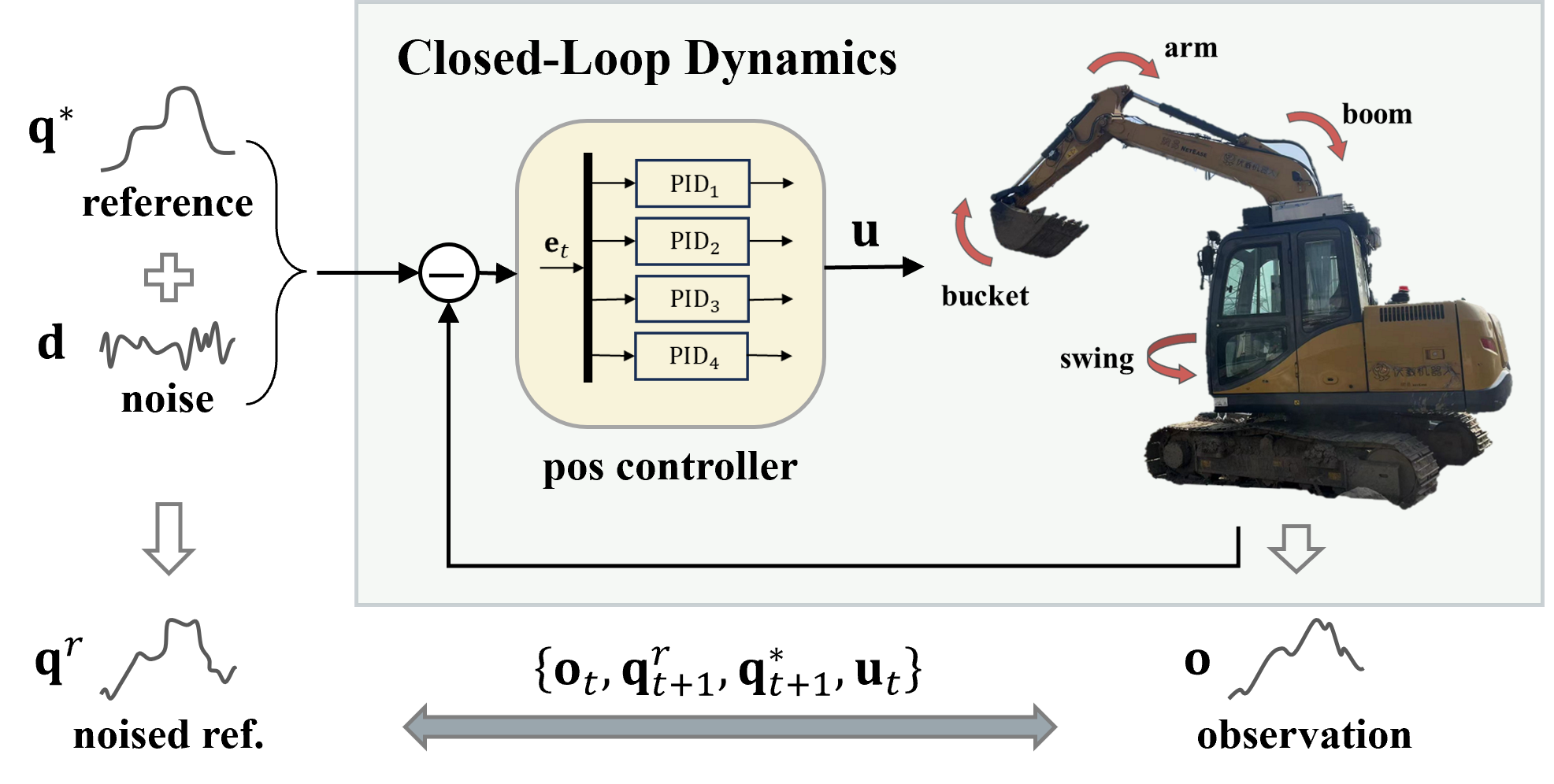}
\vspace{-8pt}
\caption{Data collection process of our method. Gaussian noise is added to the desired reference trajectories \(\mathbf{q}^*\) to generate the noisy reference trajectories \(\mathbf{q}^r\). These are tracked by the position controller on the real-world excavator to collect observation transitions for training the closed-loop dynamics model.}
\label{data_collection}
\vspace{-12pt}
\end{figure}

\subsubsection{Data Collection}
\label{data collection}
We first collect loading trajectories performed by an experienced excavator operator as the desired reference trajectories \(\mathbf{q}^*\). Then, we use a position controller to track these trajectories on the excavator, obtaining closed-loop observation transitions and the corresponding reference positions. These data constitute the training dataset for the closed-loop dynamics model.

To improve the model's robustness and generalization, as illustrated in Fig.~\ref{data_collection}, small Gaussian random noise \(\mathbf{d}_t\sim\mathcal{N}\left(0, \sigma^2\right)\) is introduced into the desired trajectory at a certain ratio during data collection:
\begin{equation}
\label{noise}
\setlength\abovedisplayskip{4pt}
\setlength\belowdisplayskip{5pt}
\mathbf{q}^r_{t+1} = \mathbf{q}^*_{t+1} + \mathbf{d}_t
\end{equation}

The trained model is expected to accurately predict observations over multiple steps and provide reliable predictions under unknown policy shifts.
We conducted an ablation study (see \ref{ablation_noise}) to further demonstrate the necessity of introducing noise into the reference trajectories during data collection.

\begin{figure}[!t]
\centering
\includegraphics[width=3.4in]{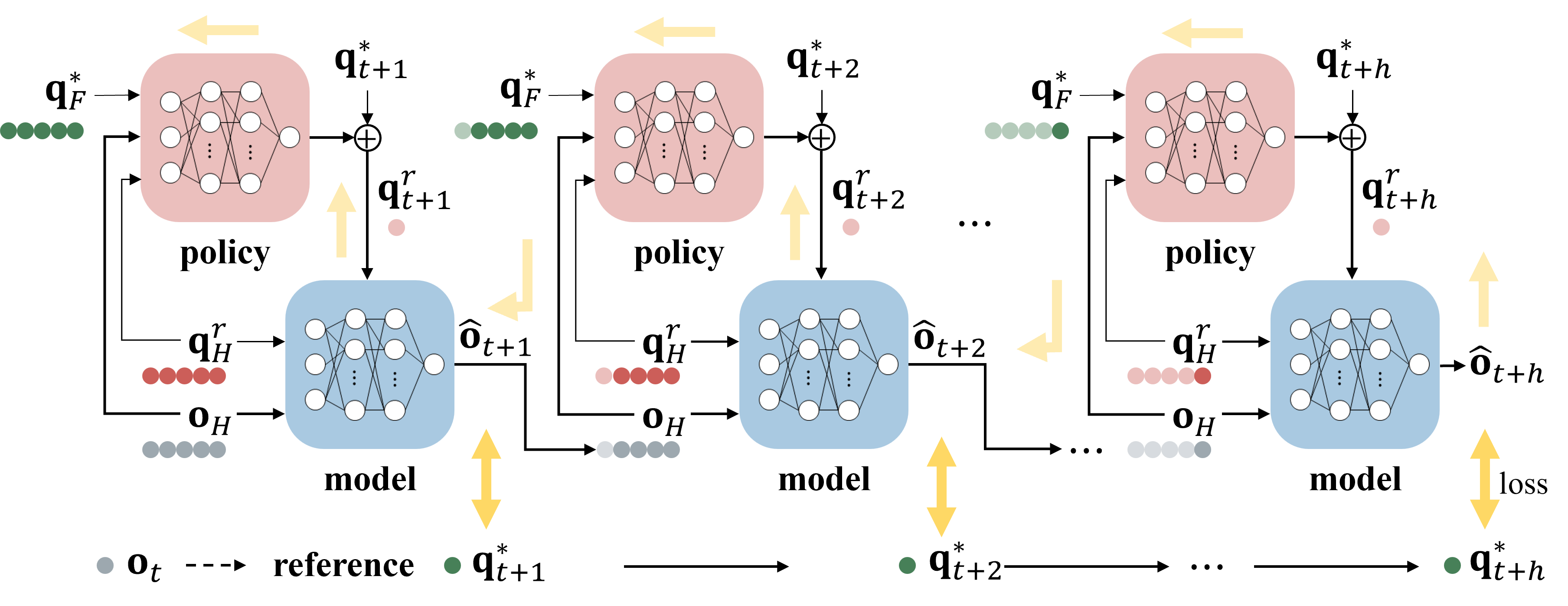}
\vspace{-18pt}
\caption{Multi-step forward and backward propagation of the trajectory adjustment policy. The black arrows indicate the forward propagation of observations over time steps, while the yellow arrows indicate the backpropagation of the tracking loss starting from step \(t + i\) to step \(t\). The policy adjusts the reference positions based on the states, and the model is driven by the reference sequence provided by the policy.}
\label{policy_learning}
\vspace{-12pt}
\end{figure}

\subsection{Trajectory Adjustment Policy}
\label{policy}
We utilize the trajectory adjustment policy \(\pi_\phi\) defined in equation (\ref{policy equation}) to improve the accuracy of the excavator's trajectory tracking by continuously modifying the reference positions for the closed-loop dynamics.

\subsubsection{Forward Propagation}
At each time step \(t\), the action \(\mathbf{a}_t\) modifies the real reference position \(\mathbf{q}^*_{t+1}\) according to equation (\ref{policy_action}), producing the updated reference \(\mathbf{q}^r_{t+1}\). This adjusted reference is then fed into the model to predict the next observation. Given the future desired positions, the policy iteratively computes the reference positions \(\mathbf{q}^r\) for the subsequent \( h \) steps. Through this process, the model predicts the observation sequence \( \hat{\mathbf{o}}_F = \left(\hat{\mathbf{o}}_{t+1}, ... \hat{\mathbf{o}}_{t+h}\right)^T \).

\subsubsection{Backward Propagation}
Consistent with the model training approach, the policy is optimized using multi-step gradient backpropagation \cite{ref20} to ensure long-term tracking accuracy. Notably, the model's parameters remain fixed and do not accumulate gradients during the policy's backpropagation. The tracking loss combines weighted prediction errors over the \( h \)-step horizon:
\begin{equation}
\setlength\abovedisplayskip{0pt}
\setlength\belowdisplayskip{3pt}
L_{track}(\phi) = \frac{1}{h} \sum_{i=1}^{h} \gamma^i\| \hat{\mathbf{q}}_{t+i} - \mathbf{q}^*_{t+i} \|^2
\end{equation}
where \(\gamma\) is the discount factor, \( \hat{\mathbf{q}}_{t+i} \) is the model-predicted position, and \( \mathbf{q}^*_{t+i} \) is the desired position at time \( t+i \).

The policy's multi-step forward and backward propagation processes are illustrated in Fig. \ref{policy_learning}.

To prevent the policy from overfitting to model prediction errors, we incorporate a regularization loss:
\begin{equation}
\setlength\abovedisplayskip{1pt}
\setlength\belowdisplayskip{3pt}
L_{reg}(\phi) = \frac{1}{h-1} \sum_{i=2}^{h} \|\mathbf{a}_{t+i} - \mathbf{a}_{t+i-1} \|^2
\end{equation}
This regularization term constrains the action differentials between consecutive steps, effectively preventing excessive responsiveness to minor prediction errors. Additionally, it enforces temporal smoothness in the trajectory adjustments, which promotes stable tracking performance throughout the operation.
We conducted an ablation study (see \ref{ablation_smooth}) to further demonstrate the performance benefits of incorporating regularization in the trajectory adjustment policy training.

\section{Experiments}
In this section, we validate our EfficientTrack method through both simulation and real-world excavator experiments. We compare our approach with existing learning-based methods to demonstrate that EfficientTrack offers high efficiency, high precision, and smooth trajectory tracking.  

\subsection{Setup}

\subsubsection{Environment}
In our real-world experiments, we employ the XCMG 7.5-ton XE75 excavator, fitted with an Inertial Measurement Unit (IMU) and a GNSS-aided Inertial Navigation System (GNSS/INS) as sensors. The IMU data is processed to derive joint angles and angular velocities for the boom, arm, and bucket, while the swing angle and angular velocity of the cab are obtained from the GNSS/INS. Control signals are sent to a Programmable Logic Controller (PLC), which adjusts the PWM duty cycle applied to the hydraulic actuators at each joint, enabling movement of the excavator.

We also built a data-driven simulation of the excavator based on real-world data. By collecting operational data under diverse conditions, we trained a neural network model that takes PWM duty cycle as input and predicts joint angular velocities and positions. This high-fidelity simulator replicates excavator dynamics and enables safe online learning experiments in a controlled environment. Compared to traditional physical models, our approach offers better adaptability to real actuator behavior, supporting more reliable deployment on actual machines.

\begin{figure}[!t]
\centering
\includegraphics[width=3.5in]{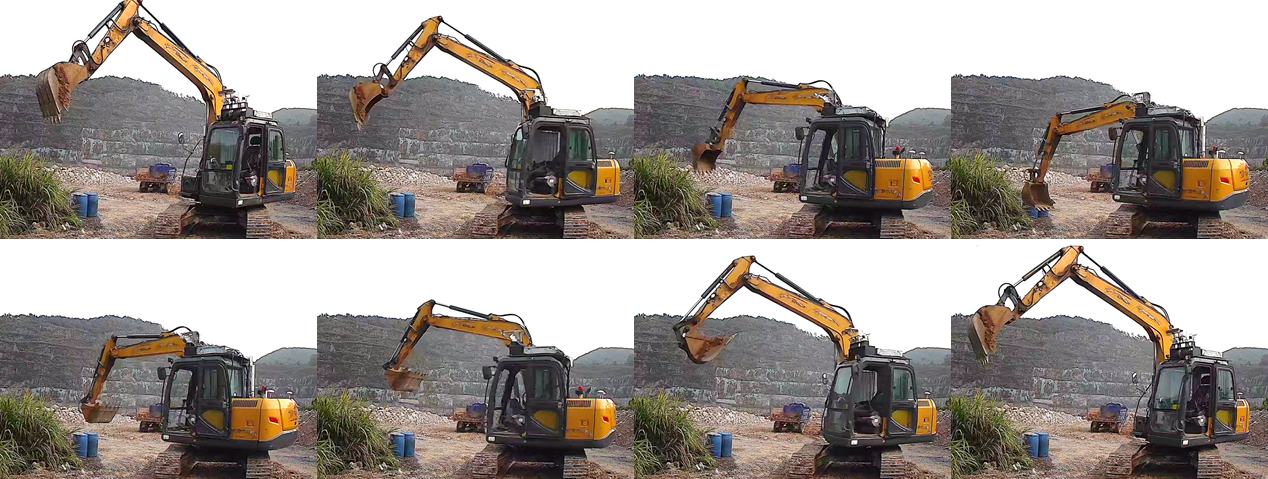}
\caption{Excavator loading process. From top left to bottom right, the steps are: bucket positioning, material loading, material transfer, and material dumping.}
\vspace{-12pt}
\label{excavator_load}
\end{figure}

We use 48 loading trajectories from experienced excavator operators as reference trajectories, each lasting around 20 seconds with a sampling frequency of 20 Hz. The complete loading process includes four main steps: bucket positioning, material loading, material transfer, and material dumping, as shown in Fig. \ref{excavator_load}.

\subsubsection{Training and Testing}
Our method consists of a training phase and an implementation phase. To verify the effectiveness and generalization of our approach, we randomly divide 48 reference trajectories into 40 for training and 8 for testing.

During training, we follow the procedure outlined in Algorithm \ref{algorithm1} for data collection, model training, and policy training. The maximum standard deviation for Gaussian noise (\(\sigma_{max}\)) is set to 0.05 radians. The number of training epochs (\(N_{epoch}\)) is 2000, and the learning horizon (\(h\)) is 20 steps. A weight decay factor (\(w\)) of 0.003 is applied to the model. The policy employ a discount factor (\(\gamma\)) of 0.98 and a smoothing factor (\(k_{smooth}\)) of 1.0. Both the model and the policy utilize a Multi-Layer Perceptron (MLP) with 6 hidden layers, each consisting of 512 units, incorporating LayerNorm \cite{ba2016layer} and ELU activation functions \cite{clevert2015fast}. The output layer employ Tanh activation to constrain the output range. Training is performed using mini-batch stochastic gradient descent with a batch size of 2048, optimized by the Adam optimizer \cite{kingma2014adam} with a learning rate of \(1\times10^{-5}\).

During testing, we implement the policy network following the procedure outlined in Algorithm \ref{algorithm2}.

\begin{algorithm}[!t]
\caption{Implementation of EfficientTrack Policy}\label{algorithm2}
\begin{algorithmic}[1]
\STATE \textsc{ImplementPolicy}(\(\mathcal{S}_{reftraj}, \pi_\phi\))

\STATE \hspace{0.4cm}\(\mathcal{S}_{record}\gets[]\)

\STATE \hspace{0.4cm}\textbf{for} \(reftraj\) \textit{\textbf{in}} \(\mathcal{S}_{reftraj}\) \textbf{do}

\STATE \hspace{0.8cm}\(\{\mathbf{q}^*_0, \mathbf{q}^*_1, ..., \mathbf{q}^*_{T}\} \gets reftraj\)

\STATE \hspace{0.8cm}\(\textsc{EnvReset}(\mathbf{q}^*_0), \mathbf{q}^r_0\gets\mathbf{q}^*_0\)

\STATE \hspace{0.8cm}\(\mathbf{o}_t\gets\textsc{GetObservation}(), \mathbf{o}_H\gets[\mathbf{o}_0, ..., \mathbf{o}_0]_h\)

\STATE \hspace{0.8cm}\(\mathbf{q}^r_H\gets[\mathbf{q}^r_0, ..., \mathbf{q}^r_0]_h, \mathbf{q}^*_F\gets[\mathbf{q}^*_1, ..., \mathbf{q}^*_h]\)

\STATE \hspace{0.8cm}\textbf{for} \(t\) \textit{\textbf{in}} \(\text{range}(T)\) \textbf{do}

\STATE \hspace{1.2cm}\(\mathbf{o}_t\gets\textsc{GetObservation}(), (\mathbf{q}_t, \mathbf{\dot{q}}_t)\gets\mathbf{o}_t\)

\STATE \hspace{1.2cm}\(\mathbf{o}_H\gets\left[\mathbf{o}_H\text{[1:]}, \mathbf{o}_t\right], \mathbf{q}^r_H\gets\left[\mathbf{q}^r_H\text{[1:]}, \mathbf{q}^r_t\right]\)

\STATE \hspace{1.2cm}\(\mathbf{q}^*_F\gets\left[\mathbf{q}^*_F\text{[1:]}, \mathbf{q}^*_{t+h}\right]\)

\STATE \hspace{1.2cm}\(\mathbf{a}_t\gets\pi_\phi(\mathbf{o}_H, \mathbf{q}^r_H, \mathbf{q}^*_F), \mathbf{q}^r_{t+1}\gets\mathbf{q}^*_{t+1}+\mathbf{a}_t\)

\STATE \hspace{1.2cm}\(\mathbf{u}_t\gets\textsc{PdController}(\mathbf{q}_t,\mathbf{q}_{t-1}, \mathbf{q}^r_{t+1},\mathbf{q}^r_t)\)

\STATE \hspace{1.2cm}\(done\gets\textsc{EnvStep}(\mathbf{u}_t)\)

\STATE \hspace{1.2cm}\(\mathcal{S}_{record}\gets\mathcal{S}_{record} \cup \{\mathbf{o}_t, \mathbf{q}^r_{t+1}, \mathbf{q}^*_{t+1}, \mathbf{u}_t, done\}\)

\STATE \hspace{0.4cm}\textbf{return} \(\mathcal{S}_{record}\)

\end{algorithmic}
\end{algorithm}

\begin{table*}[!t]
\begin{center}
\caption{Metrics of Comparative Experiments in Simulation}
\label{table1}
\begin{tabular}{llccccccccc}
\toprule   
\textbf{Env.}& \textbf{Method} & \textbf{I.S.} \textdownarrow & \textbf{I.T.} (h) \textdownarrow & \(\textbf{1}^{\textbf{st}}\)\textbf{oSMT} & \(\textbf{2}^{\textbf{nd}}\)\textbf{oSMT} & \textbf{MAE} & \textbf{RMSE} & \textbf{FMAE} & \textbf{ETD-MAE} & \textbf{ETD-FMAE}
\\ [0.5ex]
\midrule
 & PD Controller & - & - & \textbf{0.67\degree} & 0.70\degree & 2.73\degree & 3.14\degree &  2.93\degree & 14.73 (cm) & 11.25 (cm) \\[0.0ex]
\midrule
 & PPO & 12,361,696 & 171.69(/16) & 3.28\degree & 2.91\degree & 24.94\degree & 28.82\degree &  19.80\degree & 183.45 & 200.82 \\[0.5ex]
 & SAC & 486,858 & 6.76 & 1.43\degree & 1.67\degree & 2.78\degree & 3.68\degree &  5.85\degree & 34.80 & 70.82 \\[0.5ex]
Open-loop & TD-MPC & 241,468 & 3.35 & 1.35\degree & 1.69\degree & 1.66\degree & 2.18\degree &  1.15\degree & 17.33 & 12.66 \\[0.5ex]
 & SuperTrack & 543,620 & 7.55 & \textbf{0.75\degree} & 0.84\degree & 0.58\degree & \textbf{0.82\degree} &  0.46\degree & 6.74 & 5.97 \\[0.5ex]
 & \textbf{Ours (Open.)} & 176,720 & 2.45 & 0.76\degree & \textbf{0.63\degree} & \textbf{0.53\degree} & 0.92\degree & \textbf{0.45\degree} & \textbf{5.19} & \textbf{3.86} \\[0.0ex]
\midrule
 & PPO & 10,590,994 & 147.10(/16) & 0.70\degree & 0.76\degree & 4.45\degree & 5.15\degree &  5.11\degree & 61.57 & 50.47 \\[0.5ex]
 & SAC & 442,892 & 6.15 & 1.10\degree & 1.38\degree & 1.54\degree & 2.18\degree &  3.84\degree & 14.92 & 50.90 \\[0.5ex]
Closed-loop & TD-MPC & 197,890 & 2.75 & 0.94\degree & 1.13\degree & 0.60\degree & 0.80\degree &  0.49\degree & 7.99 & 5.90 \\[0.5ex]
 & SuperTrack & 523,541 & 7.27 & 0.71\degree & 0.73\degree & 0.34\degree & 0.49\degree &  0.36\degree & 3.96 & 3.64 \\[0.5ex]
 & \textbf{Ours} & 176,720 & 2.45 & \textbf{0.67\degree} & \textbf{0.59\degree} & \textbf{0.32\degree} & \textbf{0.46\degree} & \textbf{0.23\degree} & \textbf{3.94} & \textbf{2.00} \\[0.0ex]
\bottomrule
\end{tabular}
\end{center}
\vspace{-15pt}
\end{table*}

\subsubsection{Metrics}
We have the following metrics related to tracking error:
\begin{itemize}
\item \textbf{MAE}: Mean absolute tracking error between the angle position \(\mathbf{q}_t\) and the desired position \(\mathbf{q}^*_t\) for the four joints.
\item \textbf{RMSE}: Root mean square tracking error between the position \(\mathbf{q}_t\) and the desired position \(\mathbf{q}^*_t\).
\item \textbf{FMAE}: Mean absolute tracking error between the final step position \(\mathbf{q}_T\) and the desired position \(\mathbf{q}^*_T\).
\item \textbf{ETD-MAE}: Mean absolute distance error between the end effector points \(\left(x_t, y_t, z_t\right)\) and the desired points \(\left(x^*_t, y^*_t, z^*_t\right)\) using the excavator's forward kinematics. 
\item \textbf{ETD-FMAE}: Mean absolute distance error between the final step end effector points \(\left(x_T, y_T, z_T\right)\) and the desired points\(\left(x^*_T, y^*_T, z^*_T\right)\).
\end{itemize}

The following metrics are associated with the smoothness of the tracking trajectory:
\begin{itemize}
\item \(1^{st}\) Order Smoothness (\(\textbf{1}^{\textbf{st}}\)\textbf{oSMT}): Root mean square error of the position between consecutive steps.
\begin{equation}
\setlength\abovedisplayskip{4pt}
\setlength\belowdisplayskip{4pt}
\begin{small}
1^{st}\text{oSMT} = \sqrt{\frac{1}{T}\sum_{i=1}^{T}\|\mathbf{q}_i-\mathbf{q}_{i-1}\|^2}
\end{small}
\end{equation}
\item \(2^{nd}\) Order Smoothness (\(\textbf{2}^{\textbf{nd}}\)\textbf{oSMT}): Root mean square error of the change in position.
\begin{small}
\begin{equation}
\setlength\abovedisplayskip{4pt}
\setlength\belowdisplayskip{4pt}
2^{nd}\text{oSMT} =\sqrt{\frac{1}{T-1}\sum_{i=2}^{T}\|\left(\mathbf{q}_i-2\mathbf{q}_{i-1}+\mathbf{q}_{i-2}\right)\|^2}
\end{equation}
\end{small}
\end{itemize}

The following metrics indicate the data requirements of the algorithm:
\begin{itemize}
\item Interaction Step (\textbf{I.S.}): The total number of interactions between the agent and the environment when the model converges to the optimal value, or the amount of data used in the EfficientTrack training dataset.
\item Interaction Time (\textbf{I.T.}): The total interaction time with the environment, calculated by dividing the Interaction Step by the sampling frequency.
\end{itemize}

All the above metrics are obtained during testing, averaged over the testing set, with none of the test trajectories included in the training set.

\subsection{Comparative Study}

\begin{figure}[!t]
\centering
\includegraphics[width=3.2in]{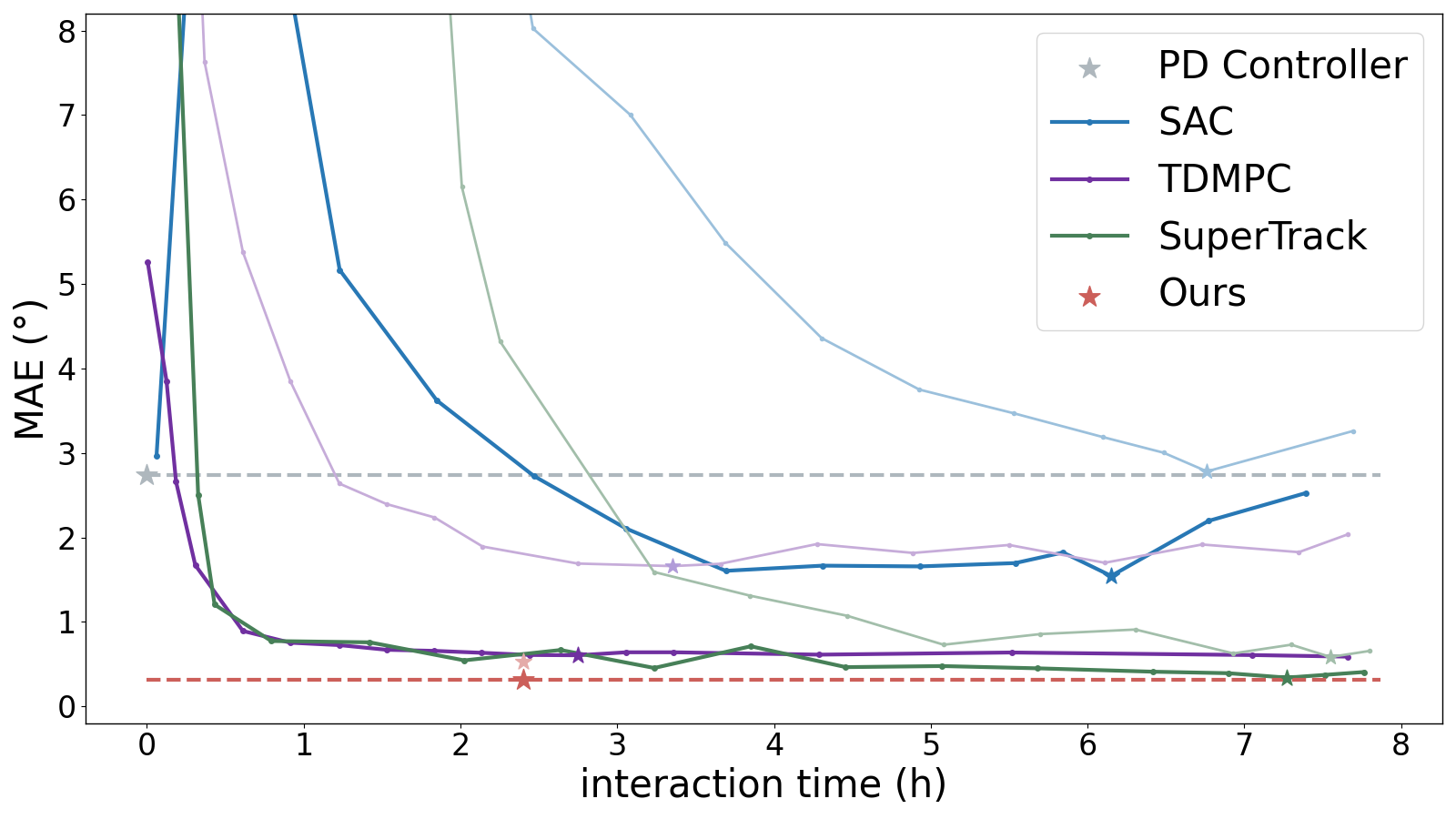}
\vspace{-5pt}
\caption{MAE variation with interaction time for different methods. The dark curves represent testing results in the closed-loop environment, while the light curves represent those in the open-loop environment. The stars indicate the points where MAE converges to its minimum. Our method achieves lower tracking errors than the baselines with the shortest interaction time.}
\label{comparative_experiment}
\vspace{-12pt}
\end{figure}

To demonstrate the learning efficiency and tracking accuracy of our method, we compare it against both widely adopted and state-of-the-art reinforcement learning algorithms for online training in trajectory tracking:
\begin{itemize}
\item \textbf{PPO} \cite{ref18}: A model-free, on-policy RL algorithm. We utilized PPO with 16-thread agents.
\item \textbf{SAC} \cite{ref19}: A model-free, off-policy RL algorithm. We configured SAC with a buffer size of 100,000.
\item \textbf{TD-MPC} \cite{ref17}: A SOTA model-based, off-policy RL algorithm for continuous action space problem.
\item \textbf{SuperTrack} \cite{ref20}: A model-based, off-policy online learning algorithm for motion tracking.
\end{itemize}

To demonstrate the advantages of introducing closed-loop dynamics, we compare all these baselines and our method executed in both an open-loop simulation environment and a closed-loop environment.
\begin{itemize}
\item \textbf{Open-loop environment}: In this case, a control policy is learned for the excavator's joints, as is commonly the case when solving trajectory tracking problems. The policy output (action) is the PWM control \(\mathbf{u}\) applied directly to the hydraulic valves of the excavator's joints. The policy inputs are historical observations \(\mathbf{o}_H\), historical actions \(\mathbf{a}_H\) and future desired positions \(\mathbf{q}^*_F\).
\item \textbf{Closed-loop environment}: In this environment, a trajectory adjustment policy is learned for trajectory tracking of the closed-loop dynamics, see \ref{policy}. A fine-tuned Proportional-Differential position controller (PD controller) is used for preliminary closed-loop tracking.
\end{itemize}


Experimental results presented in Table \ref{table1} demonstrate that the model-based approach with multi-step gradient propagation outperforms model-free methods. Furthermore, our method achieves superior convergence performance under closed-loop dynamics with the shortest interaction time among all baselines. This suggests that directly supervising with tracking error, rather than learning a value function via rewards, may be a more effective strategy for enhancing tracking precision. For more implementation details, please refer to the code.

\subsection{Ablation Study}

\subsubsection{Mixed Noise Data}
\label{ablation_noise}
In our experiments, we find that adding random Gaussian noise to a portion of the reference trajectories during data collection, while keeping another portion noise-free, results in a model with lower prediction errors and a policy with higher tracking accuracy. We experimented with noise-to-noise-free ratios (\textbf{N.R.}) of 9:1, 9:2, 9:3, and 9:4, and found that a 9:2 ratio achieved the best performance.

To demonstrate the advantage of mixed noise datasets, we conducted ablation studies by removing either the noisy data or the noise-free data. The training set consists of 40 reference trajectories totaling 17,672 steps, which is the maximum capacity for the noise-free dataset. For fair comparison, we set up experiments with equal data volumes for the noisy dataset and the mixed dataset.

Training set model prediction error (\textbf{Tr.MPE}) is defined as the mean absolute model prediction error on the training set \(\mathcal{S}_{dataset}\), whereas testing model prediction error (\textbf{Te.MPE}) refers to the same metric computed on the recorded testing data \(\mathcal{S}_{record}\) during implementation. The results, shown in Table \ref{table2}, indicate that the model trained on the mixed noise dataset achieves significantly lower model prediction error and policy tracking error during testing.

\begin{table}[!t]
\begin{center}
\caption{Metrics of Ablation Study on Noise Rate}
\label{table2}
\begin{tabular}{lccccc}
\toprule   
\textbf{N.R.} & \textbf{Tr.MPE} & \textbf{Te.MPE} & \textbf{MAE} & \textbf{RMSE} & \textbf{FMAE} \\ [0.0ex]
\midrule
No Noise & 0.91\degree & 1.34\degree & 1.62\degree & 2.19\degree & 2.44\degree \\[0.5ex]
Only Noise & 0.95\degree & 1.09\degree & 1.76\degree & 2.21\degree & 1.28\degree \\[0.5ex]
9:2 & \textbf{0.67\degree} & \textbf{0.79\degree} & \textbf{0.81\degree} & \textbf{1.05\degree} & \textbf{0.65\degree} \\[0.0ex]
\bottomrule
\end{tabular}
\end{center}
\vspace{-8pt}
\end{table}

\subsubsection{Regularization Loss} 
\label{ablation_smooth}
We also conducted experiments to assess the necessity of adding a regularization loss during policy training. As shown in Table \ref{table3}, omitting the regularization loss allows the policy to achieve the lowest mean absolute tracking error (\textbf{Po.MAE}) on the model-predicted positions during training. However, this approach leads to the policy overfitting to the model's prediction errors, resulting in reduced tracking accuracy during testing. Conversely, applying a larger regularization parameter \(k_{smooth}\) produces a smoother policy but sacrifices fine-grained tracking capability, leading to an overall increase in tracking error.

\begin{table}[!t]
\begin{center}
\caption{Metrics of Ablation Study on Regularization Loss}
\label{table3}
\begin{tabular}{cccccc}
\toprule   
\(k_{smooth}\) & \textbf{Po.MAE} & \(\textbf{1}^{\textbf{st}}\)\textbf{oSMT} & \(\textbf{2}^{\textbf{nd}}\)\textbf{oSMT} & \textbf{MAE} & \textbf{RMSE} \\ [0.5ex]
\midrule
0.0 & \textbf{0.19\degree} & 0.90\degree & 1.05\degree & 0.69\degree & 0.95\degree \\[0.5ex]
\textbf{1.0} & 0.27\degree & 0.74\degree & 0.77\degree & \textbf{0.60\degree} & \textbf{0.77\degree} \\[0.5ex]
2.0 & 0.30\degree & \textbf{0.74\degree} & \textbf{0.75\degree} & 0.63\degree & 0.80\degree \\[0.0ex]
\bottomrule
\end{tabular}
\end{center}
\vspace{-15pt}
\end{table}

\begin{table*}[!t]
\begin{center}
\caption{Metrics of Experiments on the Real-World Excavator}
\label{table4}
\begin{threeparttable}
\begin{tabular}{llccccccccc}
\toprule   
\textbf{Dataset} & \textbf{Round} & \textbf{I.S.} & \textbf{I.T.} (h) & \(\textbf{1}^{\textbf{st}}\)\textbf{oSMT} & \(\textbf{2}^{\textbf{nd}}\)\textbf{oSMT} & \textbf{MAE} & \textbf{RMSE} & \textbf{FMAE} & \textbf{ETD-MAE} & \textbf{ETD-FMAE}  
\\ [0.5ex]
\midrule
& PD Controller & - & - & \textbf{0.56\degree} & \textbf{0.17\degree} & 5.15\degree & 6.56\degree &  2.14\degree & 79.21 (cm) & 28.77 (cm) \\[0.0ex]
\midrule
& \textit{Open-Loop} \tnote{1} & 176,720 & 2.45 & 0.84\degree & 0.20\degree & 17.83\degree & 26.91\degree &  8.11\degree & 254.53 & 182.13 \\[0.5ex]
& \textit{Round 1} & 176,720 & 2.45 & 0.67\degree & 0.23\degree & 1.30\degree & 1.91\degree &  0.67\degree & 17.27 & 4.81 \\[0.5ex]
\(\mathcal{S}_{dataset}\)\ \  & \textit{Round 2} & 194,392 & \ 2.45 \textbf{+0.25} \  & 0.63\degree & 0.21\degree & 0.69\degree & 0.99\degree &  0.41\degree & 6.83 & 3.66 \\[0.5ex]
& \textit{Round 3} & 212,064 & 2.70 \textbf{+0.25} & \textbf{0.62\degree} & \textbf{0.19\degree} & \textbf{0.64\degree} & \textbf{0.94\degree} & \textbf{0.27\degree} & \textbf{6.53} & \textbf{2.79} \\[0.5ex]
& \textit{Round 3-Load} \tnote{2}\ \  & 212,064 & 2.95 \textbf{+0} & \textbf{0.62\degree} & 0.21\degree & 0.68\degree & 1.03\degree &  0.32\degree & 6.91 & 3.76 \\[0.0ex]
\midrule
& \textit{Round 1} & 17,672 & 0.25 & 0.64\degree & 0.19\degree & 1.43\degree & 2.02\degree &  0.96\degree & 14.94 & 7.43 \\[0.5ex]
\(\mathcal{S}_{fastset}\) & \textit{Round 2} & 35,344 & 0.25 \textbf{+0.25} & \textbf{0.62\degree} & \textbf{0.18\degree} & 0.85\degree & 1.29\degree &  0.49\degree & 8.48 & 7.41 \\[0.5ex]
& \textit{Round 3} & 53,016 & 0.49 \textbf{+0.25} & 0.64\degree & 0.28\degree & \textbf{0.78\degree} & \textbf{1.14\degree} &  \textbf{0.44\degree} & \textbf{7.39} & \textbf{5.79} \\[0.0ex]
\bottomrule
\end{tabular}
\begin{tablenotes}
\item[1] To avoid the risk of overturning due to the bucket pressing down on the ground, the open-loop results only include a single test trajectory.
\item[2] Due to variations in the height of the soil pile, the loaded results include only test trajectories with loads of 50\% or higher.
\end{tablenotes}
\end{threeparttable}
\end{center}
\vspace{-10pt}
\end{table*}

\begin{figure*}[!t]
\centering
\includegraphics[width=7in]{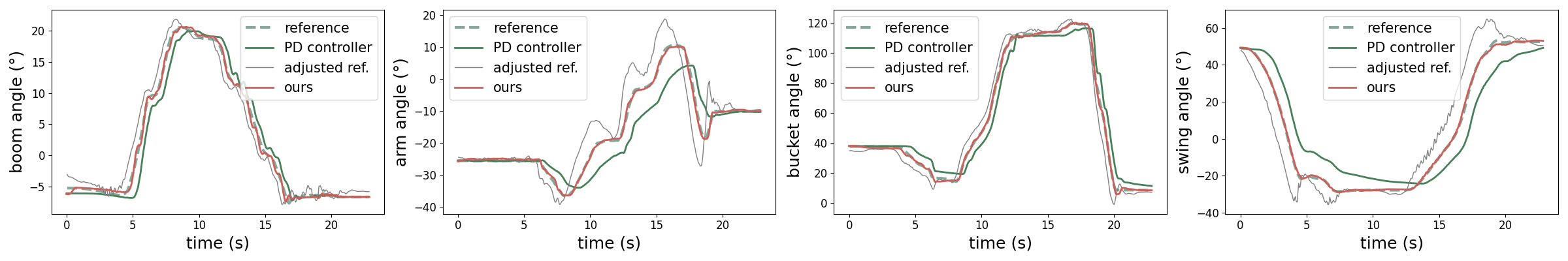}
\vspace{-22pt}
\caption{Tracking performance of the PD controller and our method on the real-world excavator. Our method (\textit{Round 3}) represents the testing outcome after the third round of training.}
\label{field_outcome}
\vspace{-12pt}
\end{figure*}

\subsection{Real-World Experiments}

\subsubsection{Continual Learning}
The training dataset \(\mathcal{S}_{dataset}\), collected under closed-loop dynamics with a PD controller (as described in Section \ref{data collection}), closely matches the original trajectory distribution. Incorporating data collected during policy implementation can help the model better capture observation transitions driven by the updated policy, resulting in a more accurate tracking performance. To verify this, we designed experiments with the following setups:
\begin{itemize}
\item \textit{Round 1}: We collected the dataset \(\mathcal{S}_{dataset}\) on the real-world excavator and trained our closed-loop dynamics model and the trajectory adjustment policy from scratch, strictly following Algorithm \ref{algorithm1}.
\item \textit{Round 2}: Using the policy \(\pi_{\phi_1}\) learned from \textit{Round 1}, we collected new data on the training set without adding noise. We then conducted a second round of training using only the newly collected data, initializing the model and policy network parameters with \(\theta_1\) and \(\phi_1\) from \textit{Round 1}.
\item \textit{Round 3}: We repeated the process by collecting data with the policy \(\pi_{\phi_2}\), again without adding noise. We trained the model and policy exclusively on this new data, initializing the parameters with \(\theta_2\) and \(\phi_2\).
\end{itemize}
As shown in Table \ref{table4}, our method in \textit{Round 1} decreases the mean absolute tracking error from 5.15\degree (achieved by the PD controller) to 1.30\degree. After further training in \textit{Round 2} and \textit{Round 3}, the tracking error decreases even more, and the smoothness improves significantly. Figure \ref{field_outcome} illustrates the performance of \(\pi_{\phi_3}\) in tracking one of the testing trajectories.

\subsubsection{Efficiency}
To demonstrate the superior efficiency of our method, we also trained our model and policy using \(\mathcal{S}_{fastset}\), which consists of training trajectories with random noise levels and has only one-tenth of the data volume of \(\mathcal{S}_{dataset}\). Remarkably, even with just 0.25 hours of interaction, our method still reduced the average angular error to approximately 1\degree, and the average end-point error to within 10 cm.

\subsubsection{Generalization}
We directly applied the policy network \(\pi_{\phi_3}\) from \textit{Round 3} to the excavator under loaded conditions without any additional training. As shown in the \textit{Round 3-Load} row, the smoothness and tracking accuracy exhibited only slight decreases. This demonstrates that our method remains effective under loaded conditions and generalizes well to different load levels.

\subsubsection{Benefits of Closed-Loop Dynamics}
To analyze the advantages of introducing closed-loop dynamics on the real excavator, we conducted an experiment using the open-loop setting for comparison. Unlike in simulation, where the policy trained under open-loop dynamics can achieve good tracking performance, the identically trained policy struggled to track the desired trajectories during the real-world experiment, as shown in the \textit{Open-Loop} row. This discrepancy is likely due to various disturbances present in the real world, such as air resistance and mechanical friction, which are not accounted for in the simulation's deterministic dynamics.
This further demonstrates the robustness of our method based on closed-loop dynamics, highlighting its practical benefits on the real machine.

\section{Conclusion}
We present EfficientTrack, a learning-based trajectory tracking method for complex nonlinear systems. Leveraging closed-loop dynamics and multi-step backpropagation, it achieves high efficiency, precision, and robustness. Extensive experiments show our method outperforms existing learning-based methods, achieving superior tracking precision and smoothness with reduced interaction time. Our method also supports continual learning and shows strong real-world applicability. Future work may extend it to different controllers, excavator types, and more complex excavation tasks, enhancing its generalizability across diverse real-world scenarios.

\bibliographystyle{IEEEtran}
\bibliography{root.bbl}

\end{document}